\documentclass[runningheads]{llncs}
\usepackage[T1]{fontenc}
\usepackage{graphicx}
\usepackage{cite}
\usepackage{amsmath,amssymb,amsfonts}
\usepackage{algorithmic}
\usepackage{graphicx}
\usepackage{textcomp}
\usepackage{xcolor}

\usepackage{listings}
\usepackage{hyperref}
\usepackage{lipsum}
\usepackage{comment}
\usepackage{booktabs}
\usepackage{microtype}

\usepackage{subcaption}  
\usepackage{tabularx}

\usepackage[export]{adjustbox}

\def\BibTeX{{\rm B\kern-.05em{\sc i\kern-.025em b}\kern-.08em
    T\kern-.1667em\lower.7ex\hbox{E}\kern-.125emX}}
    
\begin{document}

\definecolor{codegreen}{rgb}{0,0.6,0}
\definecolor{codegray}{rgb}{0.5,0.5,0.5}
\definecolor{codepurple}{rgb}{0.58,0,0.82}
\definecolor{backcolour}{rgb}{0.95,0.95,0.92}

\lstdefinestyle{mystyle}{
    backgroundcolor=\color{white},   
    commentstyle=\color{green!60!black},   
    keywordstyle=\color{blue},       
    numberstyle=\tiny\color{gray},   
    stringstyle=\color{red},          
    basicstyle=\ttfamily\tiny,  
    basicstyle=\ttfamily\fontsize{5}{5}\selectfont,  
    breaklines=true,                 
    frame=single,                     
    numbers=none,                     
    showstringspaces=false,           
    captionpos=b,                     
    morekeywords={*,def,return,TensorDef,ScalarDef,TileUsingForallOp,loop_fuse_sibling,VectorizeOp,OneShotBufferizeOp,ApplyPatternsOp,ApplyCastAwayVectorLeadingOneDimPatternsOp,ApplyTransferPermutationPatternsOp,ApplyRankReducingSubviewPatternsOp,YieldOp, forall},      
    emph={Ex,Ey,Ez,Hx,Hy,Hz,T1, NX, NY, NZ, Fx0, Fx1, Fy0, Fy1, Coef, Fo, D0, D1, tiledx, loopx, tiledy, loopy, fused_loop_result, TILE_SIZE, bodyTarget,fused_loop, tensor},         
    emphstyle=\color{purple}          
}
    
\lstset{style=mystyle}

\title{Optimizing FDTD Solvers for Electromagnetics: A Compiler-Guided Approach with High-Level Tensor Abstractions}
\titlerunning{A Compiler-Guided Approach for FDTD Solvers}
%

\author{Yifei He \and
Måns I. Andersson
\and
Stefano Markidis }
\authorrunning{Y. He et al.}
%
\institute{KTH Royal Institute of Technology\\
\email{\{yifeihe, mansande, markidis\}@kth.se}}

\maketitle              
\begin{abstract}
The Finite Difference Time Domain (FDTD) method is a widely used numerical technique for solving Maxwell's equations, particularly in computational electromagnetics and photonics. It enables accurate modeling of wave propagation in complex media and structures but comes with significant computational challenges. Traditional FDTD implementations rely on handwritten, platform-specific code that optimizes certain kernels while underperforming in others. The lack of portability increases development overhead and creates performance bottlenecks, limiting scalability across modern hardware architectures. To address these challenges, we introduce an end-to-end domain-specific compiler based on the MLIR/LLVM infrastructure for FDTD simulations. Our approach generates efficient and portable code optimized for diverse hardware platforms.We implement the three-dimensional FDTD kernel as operations on a 3D tensor abstraction with explicit computational semantics. High-level optimizations such as loop tiling, fusion, and vectorization are automatically applied by the compiler. We evaluate our customized code generation pipeline on Intel, AMD, and ARM platforms, achieving up to $10\times$ speedup over baseline Python implementation using NumPy.

\end{abstract}

\section{Introduction}

\begin{figure}[ht]
    \centering
    \begin{subfigure}[t]{0.48\textwidth}
        \centering
        \includegraphics[width=\linewidth]{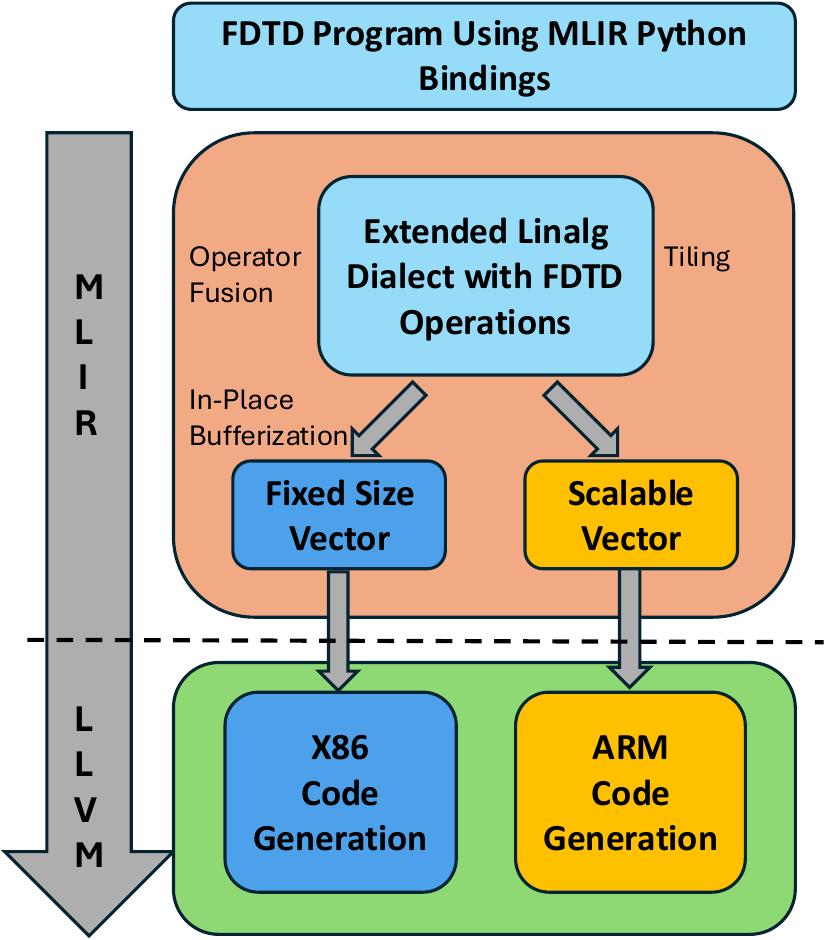}
        \caption{Overview of the FDTD Domain-Specific Compiler}
        \label{fig:fdtd_overview}
    \end{subfigure}\hfill
    \begin{subfigure}[t]{0.48\textwidth}
        \centering
        \includegraphics[width=\linewidth]{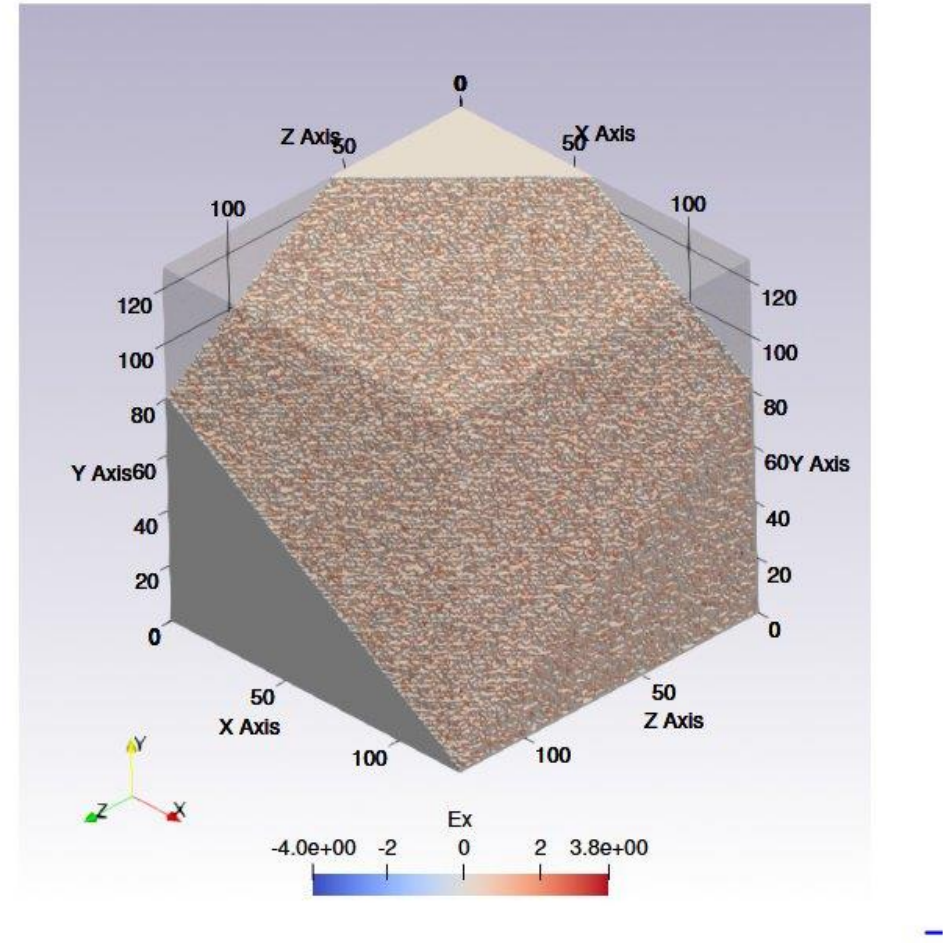}
        \caption{Visualization of cavity with ParaView}
        \label{fig:ex2}
    \end{subfigure}
    \caption{(a): Diagram illustrating the overall domain-specific compilation and code-generation pipeline for FDTD. (b): The $E_x$ field in a cavity domain with a random initial state, showing only positive values after thresholding. Note the enforced PEC boundary conditions.}
    \label{fig:fdtd_comparison}
\end{figure}

The Finite-Difference Time-Domain (FDTD) method~\cite{taflove2005computational,1138693} is one of the most used numerical algorithms for solving Maxwell’s equations, the governing equations describing electromagnetic wave propagation in a vacuum or a medium. FDTD is applied to a range of different important electromagnetic applications, ranging from antenna design~\cite{hanson2021broadband} to radar cross-section analysis~\cite{taflove1989review}, optical device modeling~\cite{taflove2013advances}, and plasma modeling~\cite{birdsall2018plasma}. For instance, FDTD methods are widely in use to model and design 6G wireless technologies~\cite{papadopoulos2022open}. By discretizing both time and spatial domains, FDTD enables accurate simulations of complex electromagnetic environments. However, the FDTD method relies on its ability to perform accurate, high-resolution simulations, often requiring millions of grid cells and fine-grained time steps to capture the small-scale features of the electromagnetic fields accurately and resolve fundamental numerical stability, such as the Courant-Friedrichs-Lax (CFL) condition~\cite{bondeson2012computational}. Figure~\ref{fig:ex2} shows a ParaView visualization of the box cavity simulation used for performance benchmarking.

The computational requirements of FDTD are huge due to the high spatial and temporal resolutions required for accurate electromagnetic simulations and the need to simulate relatively large systems for an extended period while still satisfying the CFL condition. As the method involves updating the electromagnetic field components in every cell at each time step, FDTD naive implementations tend to be memory-bound and exhibit relatively low computational density~\cite{hadi2012case}.


Thus, optimizing memory utilization, maximizing cache efficiency, and exploiting parallelism is fundamental to achieving scalable, high-performance Finite-Difference Time-Domain (FDTD) simulations. Conventional FDTD implementations predominantly rely on manually optimized, hardware-specific computational kernels. Although these kernels achieve high performance on select platforms and problem configurations through labor-intensive tuning, they consistently underperform on others due to platform and case-specific constraints. This fragmented, non-portable methodology introduces inherent trade-offs: platform-centric optimizations compromise maintainability and restrict scalability, while recurring adaptation to evolving hardware paradigms perpetuates development bottlenecks, stifling productivity in large-scale electromagnetic modeling.

To overcome these limitations, we introduce an end-to-end domain-specific compiler for FDTD simulations, built on the MLIR~\cite{lattner2021mlir}/LLVM~\cite{lattner2004llvm} infrastructure. MLIR is an extensible compiler infrastructure~\cite{lattner2021mlir} originally designed to address emerging computational workloads, particularly in machine learning and deep learning. Developed initially at Google, MLIR is now an integral part of the LLVM ecosystem. Its deep integration with LLVM allows it to leverage LLVM’s backend for code generation while introducing new abstractions and transformations that surpass traditional single-layer compilers in performance optimizations. As HPC workloads increasingly converge with AI-driven computing—sharing common optimization patterns such as tensor computations, data reuse, and parallel execution—MLIR’s capabilities can be effectively extended to domains beyond AI. 

Our framework utilizes declarative, tensor-based abstractions with FDTD-specific operators that encode domain semantics, including computational workloads, iteration space boundaries, and memory access patterns. By decoupling high-level physical modeling from low-level hardware constraints, our approach eliminates the need for manual optimization of nested loops, data layouts, or architecture-specific parallelism. Users simply define the computational workload dimensions, while the compiler autonomously applies hardware-aware optimizations—such as loop tiling, fusion, vectorization, and parallelization—tailored to the target architecture. We demonstrate the efficiency of our domain-specific compiler across multiple platforms, including Intel, AMD, and ARM CPUs. An overview of our compiler framework is presented in Figure~\ref{fig:fdtd_overview}. More specifically, this work makes the following contributions:
\vspace{-2pt} 
\begin{itemize}
    \item FDTD-specific kernels are implemented using high-level tensor abstractions, enabling automatic optimizations such as tiling and fusion by leveraging tensor expressions and domain-specific FDTD knowledge.

    \item Automated extraction of hardware-specific parallelism, integrating vectorization and architecture-aware code generation for Intel, AMD, and ARM CPUs through a unified MLIR/LLVM backend.

    \item Performance evaluation and analysis of our end-to-end domain-specific compiler for the FDTD solver on Intel, AMD, and ARM CPUs, achieving up to $10\times$ speedup over the baseline NumPy implementation.


\end{itemize}

\section{Design and Implementation of the FDTD Solver Using MLIR}
This section presents the mathematical formulation of the FDTD method, the baseline NumPy implementation, and its high-level tensor representation in our domain-specific compiler.

\subsubsection{Mathematical Formulation of the FDTD Algorithm}
The FDTD method is the most commonly used method for numerical simulation of Maxwell's equations \cite{bondeson2012computational}. The FDTD method solves two curl equations coming from Ampere's and Faraday's laws in SI:
\begin{equation}
\begin{aligned}
\frac{\partial \mathbf{E}}{\partial t} &= \frac{1}{\epsilon} \left( \nabla \times \mathbf{H} - \mathbf{J} \right), \quad 
\frac{\partial \mathbf{H}}{\partial t} &= \frac{1}{\mu} \nabla \times \mathbf{E}
\end{aligned}
\end{equation}
where, in a 3D Cartesian geometry, $\mathbf{E} = (E_x, E_y, E_z)$ is the electric field vector, $\mathbf{H} = (H_x, H_y, H_z)$ is the magnetic field vector, $\mathbf{J}$ is the current density vector, $\epsilon$ is the  medium dielectric, and  $\mu$ is the permeability of the medium. The Yee method is \emph{staggered} both in space and time: the electric field components $\mathbf{E} = (E_x, E_y, E_z)$ are defined at the edges of the computational cells, and the magnetic field components $\mathbf{H} = (H_x, H_y, H_z)$ are defined at the faces of the cells. For the electric field components, assuming that there is no external current $\mathbf{J}$, the discretized update equations are:
\begin{equation}
\begin{aligned}
E_x^{n+1/2} &\approx E_x^n + \frac{\Delta t}{\epsilon} \left( \frac{\partial H_z}{\partial y} - \frac{\partial H_y}{\partial z} \right), \quad  H_x^{n+1} \approx H_x^n - \frac{\Delta t}{\mu} \left( \frac{\partial E_z}{\partial y} - \frac{\partial E_y}{\partial z} \right) \\
E_y^{n+1/2} &\approx E_y^n + \frac{\Delta t}{\epsilon} \left( \frac{\partial H_x}{\partial z} - \frac{\partial H_z}{\partial x} \right), \quad H_y^{n+1} \approx H_y^n - \frac{\Delta t}{\mu} \left( \frac{\partial E_x}{\partial z} - \frac{\partial E_z}{\partial x} \right) \\
E_z^{n+1/2} &\approx E_z^n + \frac{\Delta t}{\epsilon} \left( \frac{\partial H_y}{\partial x} - \frac{\partial H_x}{\partial y} \right), \quad H_z^{n+1} \approx H_z^n - \frac{\Delta t}{\mu} \left( \frac{\partial E_y}{\partial x} - \frac{\partial E_x}{\partial y} \right)
\end{aligned}
\end{equation}
where $E_x$, $E_y$, and $E_z$ are the electric field components at time step $n$ and at the half-time step $n + 1/2$. Similarly for the magnetic field components where $H_x$, $H_y$, and $H_z$ are the magnetic field components at time step $n$ and at time step $n + 1$.
The discretization uses first-order finite difference evaluation on staggered grids for the electric and magnetic field components. The spatial derivatives of the electric field \( \mathbf{E} = (E_x, E_y, E_z) \) and the magnetic field \( \mathbf{H} = (H_x, H_y, H_z) \) are computed using central differences and respecting the staggered grid arrangement. 

\subsection{Modeling the FDTD Simulation with Python \&  Domain-Specific Tensor Expression}
The FDTD algorithm requires the computation of curl operations for both the electric field components ($\mathbf{E}$) and the magnetic field components ($\mathbf{H}$) along each spatial direction. 
In a typical implementation, the algorithm operates within a structured loop. The computation begins with the curl of the magnetic field components, where $\mathbf{H}$ is updated based on the spatial derivatives of $\mathbf{E}$. Once the magnetic field update is complete, the curl of the electric field components is computed, updating $\mathbf{E}$ based on the spatial derivatives of $\mathbf{H}$. Boundary updates are performed after each field update step to account for the edge effects introduced by these computations. This sequential structure first updating $\mathbf{H}$, then $\mathbf{E}$, ensures that the coupled nature of Maxwell's equations is respected, providing numerical stability and accuracy. Moreover, the algorithm is inherently parallelizable, making it well suited for high performance computing (HPC) applications, where large-scale simulations of complex electromagnetic phenomena are often required.

\begin{figure*}[h!]
    \centering
    \begin{minipage}{0.48\textwidth}
        \lstset{language=Python, style=mystyle}
        \begin{lstlisting}[caption={Full FDTD algorithm in Python}, label={lst:FDTD}]
while (t < T):  # Loop over time steps
    # Compute curl for H components
    curl_Hx(Hx,Hy,Hz,Ex,Ey,Ez)    
    curl_Hy(Hx,Hy,Hz,Ex,Ey,Ez)  
    curl_Hz(Hx,Hy,Hz,Ex,Ey,Ez)  
    # Apply boundary conditions for H-field
    handle_H_edge(Hx,Hy,Hz,Ex,Ey,Ez) 
    
    # Compute curl for E components
    curl_Ex(Hx,Hy,Hz,Ex,Ey,Ez)  
    curl_Ey(Hx,Hy,Hz,Ex,Ey,Ez)  
    curl_Ez(Hx,Hy,Hz,Ex,Ey,Ez)
    # Apply boundary conditions for E-field
    handle_E_edge(Hx,Hy,Hz,Ex,Ey,Ez)   
    
    # Update time step  
    t += dt  
        \end{lstlisting}
    \end{minipage}
    \hfill
    \begin{minipage}{0.48\textwidth}
        \lstset{language=Python, style=mystyle}
        \begin{lstlisting}[caption={Naive Python: curl of Hx}, label={lst:curl}]
def curl_Hx(Hx, Ey, Ez): # Compute curl  
  for i in range(Nx):  # Loop over x 
    for j in range(Ny):  # Loop over y 
      for k in range(Nz):  # Loop over z 
        # Update Hx 
        Hx[i, j, k] += (dt / mu0) * (
            (Ey[i, j, k+1] - Ey[i, j, k]) / Dz  
          - (Ez[i, j+1, k] - Ez[i, j, k]) / Dy)  
        \end{lstlisting}
        \lstset{language=Python, style=mystyle}
        \begin{lstlisting}[caption={NumPy-based: curl of Hx}, label={lst:curl_slicing}]
def curl_slice_Hx(Hx, Ey, Ez):  
    # Update Hx with NumPy slicing
    Hx[:,:,:] += (dt / mu0) * (
        (Ey[:,:,1:] - Ey[:,:,:-1]) / Dz  \
      - (Ez[:,1:,:] - Ez[:,:-1,:]) / Dy) 
        \end{lstlisting}
    \end{minipage}

    \caption{Comparison of the full FDTD algorithm with naive and NumPy-based implementations of the curl operator for Hx.}
    \label{fig:FDTD_curl_comparison}
\end{figure*}

    
The core computational steps in the FDTD method involve repeatedly applying the curl operator to the electric and magnetic field components. Listing~\ref{lst:curl} presents a simplified Python implementation of the curl operator, showing the update for one of the magnetic field components, $ H_x$, using the electric field components $E_y$ and $E_z$. This operation is applied to each component, in turn, iterating over the grid to update the corresponding field values.
To increase the performance and readability of the Python code, it is possible to use a matrix formulation of the three nested loop iterating across the three dimensions, using NumPy features, such as slicing, as shown in Listing~\ref{lst:curl_slicing}. Such implementation, based on Python NumPy and slicing, provides higher performance as it allows for Numpy vectorization.

Our FDTD domain-specific compiler leverages MLIR’s \texttt{Linalg} dialect, which offers structured abstractions for tensor and buffer operations. To model electromagnetic simulations, we introduce a unified \texttt{curl\_step} operation that encapsulates curl computations for both $\mathbf{H}$ and $\mathbf{E}$ fields. This operation enables reuse across field types by varying input tensors while maintaining a single codebase.This operator extends the \texttt{Linalg} dialect and is defined using \texttt{Linalg}'s OpDSL, allowing declarative specification of arithmetic computations, memory access patterns, and input/output tensor dimensions. The compiler automatically infers the loop iteration space from tensor inputs and outputs, lowering them into efficient loop constructs. Critically, all loop iterators in the \texttt{curl\_step} operation are parallel, ensuring full independence between iterations. Unlike operations involving inherently sequential semantics (e.g., reductions in summation or matrix multiplication, where data dependencies arise between iterations), our formulation guarantees no such dependencies exist. This property enables aggressive compiler optimizations, including loop reordering, tiling, unrolling, and vectorization, without introducing data dependency hazards.

\begin{figure}[ht]
    \centering
    \includegraphics[width=0.8\textwidth]{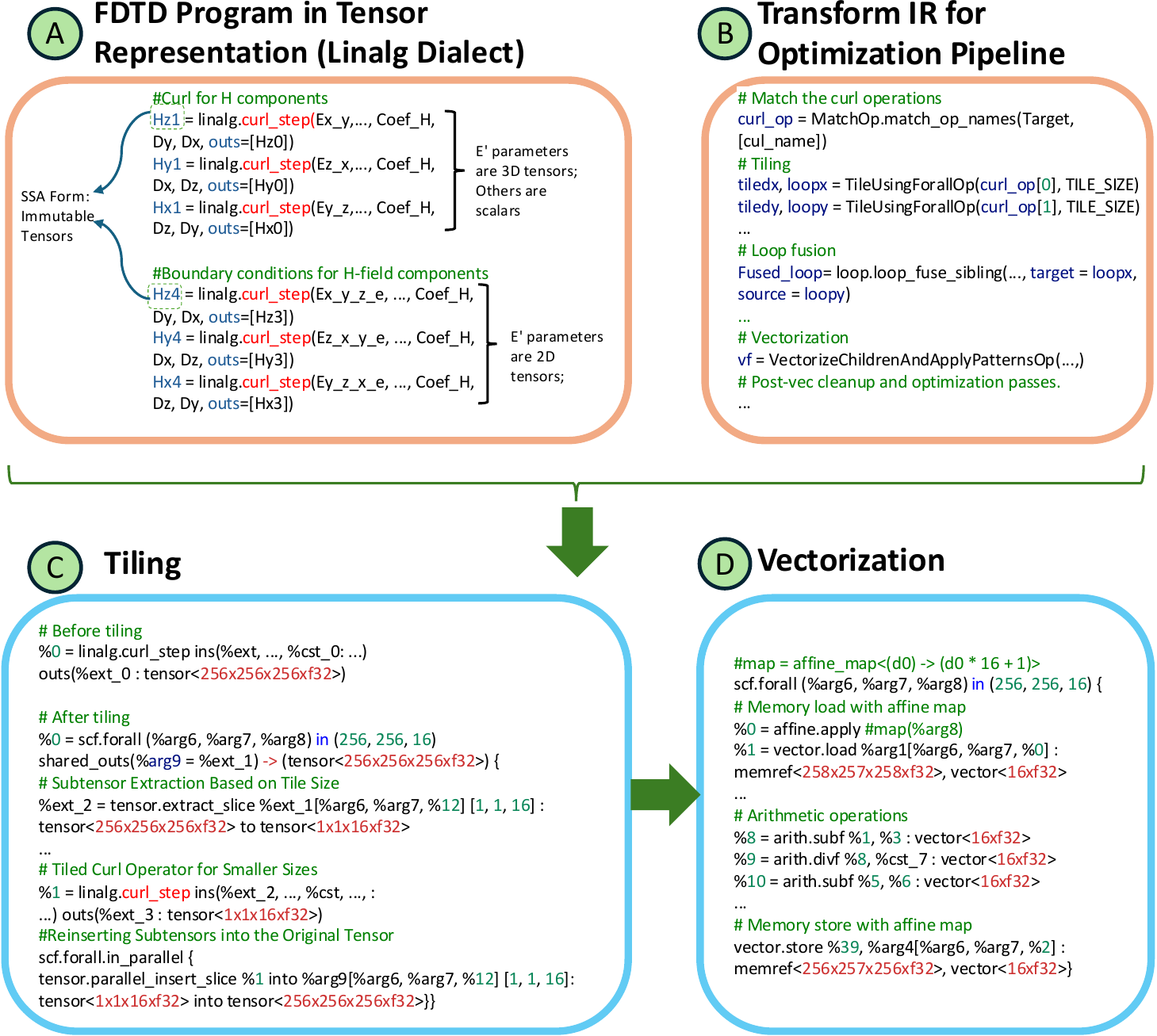} 
    \caption{MLIR example code throughout the optimization pipeline (abridged for clarity and space). (A–B): Input tensor payload IR and transform IR. (C–D): IR illustrating transformations; (C) before bufferization, (D) after bufferization, showing tensors replaced with memrefs.}
    \label{fig:Codes}
\vspace{-20pt}
\end{figure}

Figure~\ref{fig:Codes}\textbf{A} illustrates a tensor representation of the \texttt{curl\_step} operator for $\mathbf{H}$ and $\mathbf{E}$ components, along with boundary conditions using different tensor inputs and outputs. We leverage MLIR's Python bindings to directly generate FDTD-specific operations, enabling the use of Python's rich language features and seamless integration with existing Python codebases. The main computation kernels take 3D tensors as input, along with scalar values, while boundary conditions use 2D tensor inputs. The tensor representation follows the Static Single Assignment (\texttt{SSA}) form, an intermediate representation where each variable is assigned exactly once. \texttt{SSA} simplifies data-flow analysis, facilitates optimizations such as constant propagation and dead code elimination, and eliminates ambiguity in value dependencies. Consequently, each tensor appears only once in the representation. When an update is required, a new tensor is generated instead of modifying the existing one. However, during bufferization, when tensors are lowered to \texttt{memref} buffers, we enforce in-place computation to eliminate unnecessary buffer allocations, optimizing memory usage.

\section{Compilation Pipeline: Progressive Lowering and Optimization with MLIR/LLVM}
We implement an end-to-end compilation framework using MLIR/LLVM to progressively lower high-level tensor abstractions to machine code. This multi-layered framework leverages MLIR’s dialect hierarchy (Tensor → Memref → LLVM IR) to preserve high-level FDTD-specific semantics during domain-specific optimizations (e.g., loop tiling, kernel fusion, and boundary condition specialization) while progressively lowering to platform-specific code. Late-stage transformations target architecture-dependent features such as SIMD vectorization (e.g., ARM SVE intrinsics) and register allocation.

For high-level tensor abstractions, we use MLIR's \texttt{Transform} Dialect instead of traditional compiler passes. It provides fine-grained control by allowing precise transformations on specific operations and regions, unlike monolithic passes that operate on entire functions or modules. Figure~\ref{fig:Codes}:\textbf{B} shows an example code where specific \texttt{curl\_step} operations are identified and matched, followed by tiling with hyperparameters and customized loop fusion and vectorization.
\vspace{-10pt}
\subsubsection{Tiling of Tensor Operators}
Tiling is a loop transformation that partitions computations into smaller blocks, enhancing data locality, cache efficiency, and parallelism. Traditional general-purpose compilers apply tiling at low-level IR (e.g., LLVM IR), requiring exhaustive and error-prone loop bound and dependency analysis. With tensor abstractions and Linalg-structured FDTD operations, iteration spaces and memory patterns are explicitly defined in the semantics, enabling automatic partitioning of iteration spaces and data access patterns. As shown in Figure~\ref{fig:Codes}:\textbf{C}, \texttt{curr\_step} is tiled into smaller operations, with the tile size set to match the SIMD vector length.
\vspace{-10pt}
\subsubsection{Fusion of Tensor Operators}
After tiling, parallel \texttt{scf.forall} loops are generated over smaller \texttt{curl\_step} operations. We apply \texttt{loop\_fuse\_sibling}, which requires independent loops with matching structures. By leveraging the returned values from the tiling process, we selectively fuse specific loops, as shown in Figure~\ref{fig:Codes}:\textbf{B}, where \texttt{loopx} is fused into \texttt{loopy}. The \texttt{curl\_step} operations for the $\mathbf{H}$ component in three directions are fused into one, along with those for boundary conditions and the $\mathbf{E}$ field.

\vspace{-10pt}
\subsubsection{Bufferization: In-Place Computation}
MLIR bufferization transforms high-level tensor operations into explicit memory operations on the \texttt{memref} dialect—a hardware-aligned abstraction for low-level memory buffers. While we leverage the One-Shot Bufferization pass to automatically enforce in-place computation and minimize redundant buffer allocations/copies,it still has certain limitations.We meticulously structured the SSA tensor parameters for the \texttt{curl\_step} operations to align with the pattern constraints of MLIR’s One-Shot Bufferization pass, ultimately achieving in-place computation for the entire program. Additionally, we scheduled the vectorization after bufferization, as early vectorization can make the code more fragile and harder to analyze.
\vspace{-10pt}
\subsubsection{Fixed-Size Vectorization for x86 CPUs}
Following tiling, the partitioned \texttt{curl\_step} operations are aligned with the target SIMD vector length. We apply the \texttt{VectorizeChildrenAndApplyPatternsOp} pass to systematically vectorize the entire program, as shown in Figure~\ref{fig:Codes}:\textbf{(D)}. The innermost \texttt{curl\_step} operations are lowered to hardware-agnostic vector primitives (e.g., \texttt{vector.load}, \texttt{vector.store}, and \texttt{arith} operations on vector), decoupling algorithm semantics from architecture-specific SIMD intrinsics. These virtual vector operations abstract memory access and arithmetic logic, enabling flexibility across instruction sets (e.g., AVX2, AVX-512, or ARM SVE) without mandating early commitment to hardware specifics. During subsequent lowering stages, MLIR/LLVM’s code generation pipeline progressively transforms platform-agnostic vectors into architecture-specific SIMD instructions, as shown in Figure~\ref{fig:fdtd_overview}. Distinct code generation paths are employed for x86and ARM Scalable Vector Extension (SVE).

\subsubsection{Scalable Vectorization for ARM CPUs (SVE)}
ARM Scalable Vector Extension (SVE) is a variable-length SIMD architecture that abstracts hardware-specific vector lengths through predicated operations and gather/scatter memory semantics, enabling performance-portable vectorization across ARM processors (e.g., Neoverse V1, A64FX) without requiring code specialization for register widths. With fixed-size vectorization, the compiler can generate vector code for the A64Fx CPU, but it is limited to 128-bit NEON instructions, underutilizing the hardware's full capacity. By implementing scalable tiling and vectorization alongside specific LLVM compiler flags, we enable the generation of 512-bit SVE instructions on A64Fx.

\subsubsection{Vectorization and Post-Vectorization Optimizations} Finally, a series of operations from the \texttt{Vector} dialect refine the vectorized computation. The \texttt{ApplyRankReducingSubviewPatternsOp} reduces the rank of the tensor, focusing on smaller subviews of the tensor for better computational efficiency. The \texttt{ApplyTransferPermutationPatternsOp} optimizes memory transfer patterns, ensuring better data locality and improved memory access. Figure~\ref{fig:Codes}:\textbf{D} showcases the MLIR vectorized code, which avoids redundant for-loops generated by tiling. Furthermore, the memory access pattern is streamlined, consisting almost entirely of consecutive loads.


\section{Experimental Setup}

The primary goal is to optimize the main loop of the FDTD kernels. To evaluate the optimization, we test the kernels on a simple 3D cavity with PEC boundary conditions, varying discretization sizes and domain shapes. We measure performance using single-threaded versions for cubic domains with discretization sizes $N = 16^3, 32^3, 64^3, 128^3$. Each configuration is executed for 1,000 iterations to measure execution time, repeating the process 10 times to compute the mean and standard deviation. All tests are conducted using both single- and double-precision floating-point arithmetic.

\vspace{-10pt}
\subsubsection{Software and Hardware Setup}
The LLVM version used to embed our compilation framework and generate MLIR/LLVM code was forked from the main LLVM branch as of November 5, 2024. We configured the CMake build system to target both ARM and x86 architectures and enabled the MLIR Python bindings.

The test systems used include the General Purpose Partition (GPP) of MareNostrum 5 at Barcelona Supercomputing Center (BSC) featuring Intel Sapphire Rapids CPUs, the Dardel system at the PDC supercomputing center equipped with AMD EPYC Zen2 CPUs, and the Deucalion system with ARM A64FX CPUs. Further details are provided in Table~\ref{tab:HW}.

\begin{table}[ht]
\vspace{-20pt}  

\caption{The test-systems and hardware specification.}
\begin{center}
    \begin{tabular}{|c|c|c|c|}
\hline
System & Dardel (CPU) & MareNostrum 5 & Deucalion\\
\hline
CPU & AMD EPYC 7742 (Zen2) & Intel Sapphire Rapids 8480+ & ARM A64FX\\
\#Cores & 2 x 64 & 2 x 56 & 1 x 48 \\ 
Base clock &  2.25 GHz & 16 x 2.10 (40 x 1.70) GHz & 2.0 GHz\\
Boost & 3.40 GHz & 3.80 GHz & No \\
Memory & 512 GB & 256 GB & 32 GB \\
\hline
L1 & 96 KB / core & 70 KB / core & 64 KB / core \\
L2 & 512 KB / core & 2 MB / core & 8 MB / core\\ 
L3 & 256 MB & 105 MB (1.575 MB / core) & No L3 cache \\
\hline
ISA & SSE4, AVX256, & SSE4, AVX512, & ASIMD \\
& AVX2 FMA & AVX512 FMA, AMX & SVE\\
\hline
\end{tabular}
\label{tab:HW}
\end{center}
\vspace{-30pt}  
\end{table}

\subsubsection{Performance Metrics \& Baselines}
The performance of the MLIR FDTD solver is compared with a NumPy FDTD implementation in Listing~\ref{lst:curl_slicing} using NumPy slicing and is amenable to automatic vectorization and other optimization, enabled by using vector operation instead of nested loop. 



\section{Results}

\begin{figure*}[ht]
\vspace{-15pt}  
    \centering
    \captionsetup{justification=centering}  
    
    \begin{subfigure}[t]{0.48\textwidth}
        \centering
        \includegraphics[width=\textwidth]{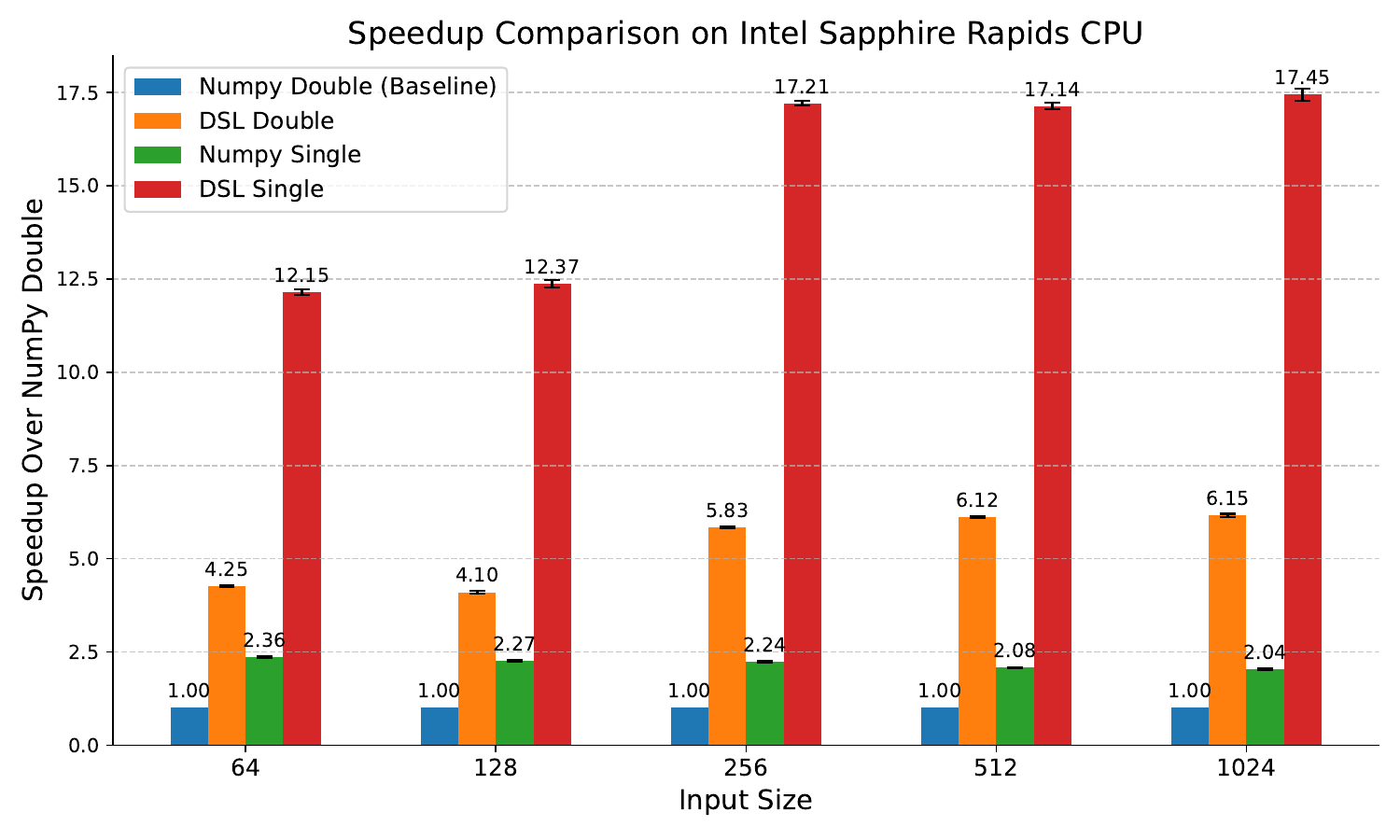}
        \caption{DSL vs. NumPy: Performance Scaling on Intel CPU.}
        \label{fig:PerfA}
    \end{subfigure}
    \hfill
    \begin{subfigure}[t]{0.48\textwidth}
        \centering
        \includegraphics[width=\textwidth]{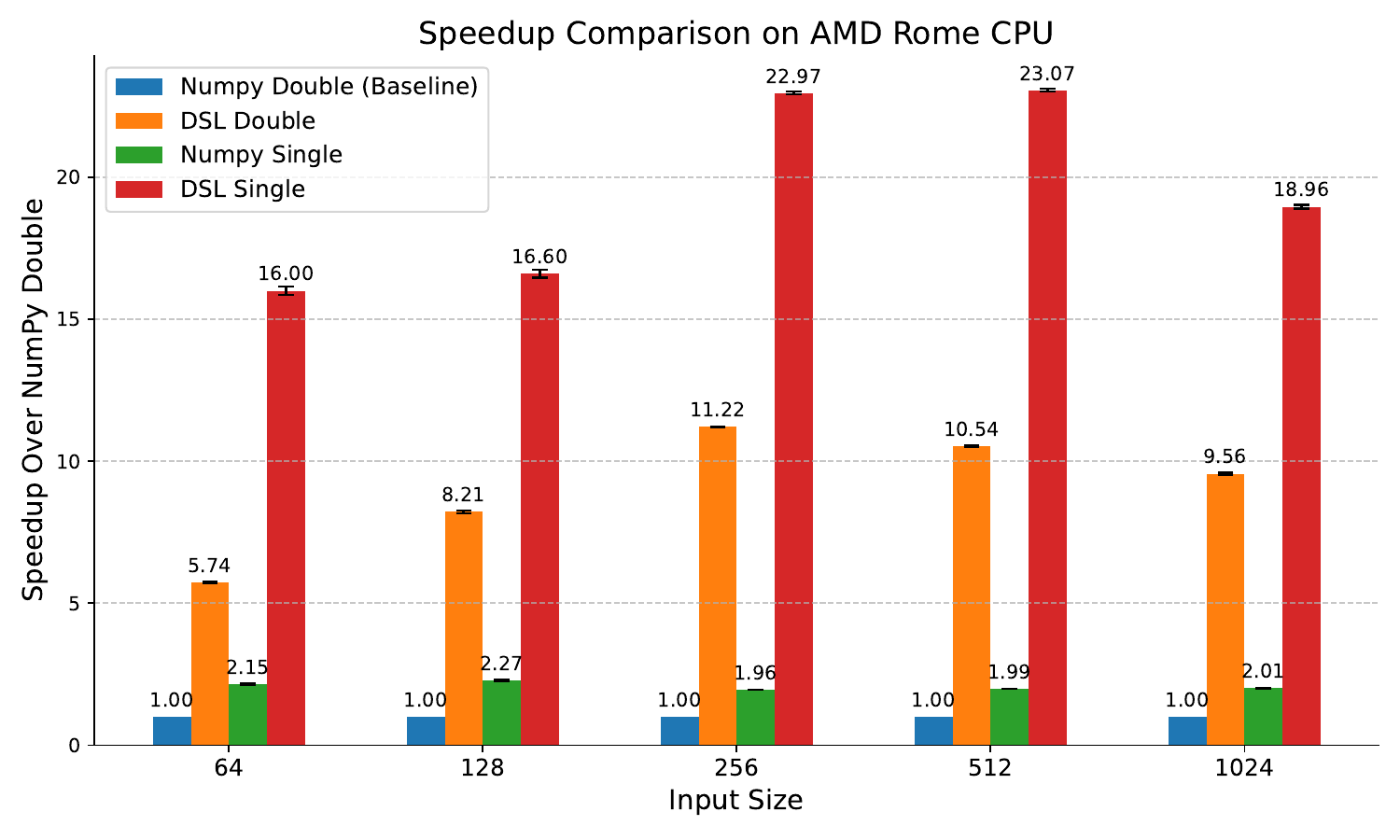}
        \caption{DSL vs. NumPy: Performance Scaling on AMD CPU.}
        \label{fig:PerfB}
    \end{subfigure}

    \vspace{5pt}  

    \begin{subfigure}[t]{0.48\textwidth}
        \centering
        \includegraphics[width=\textwidth]{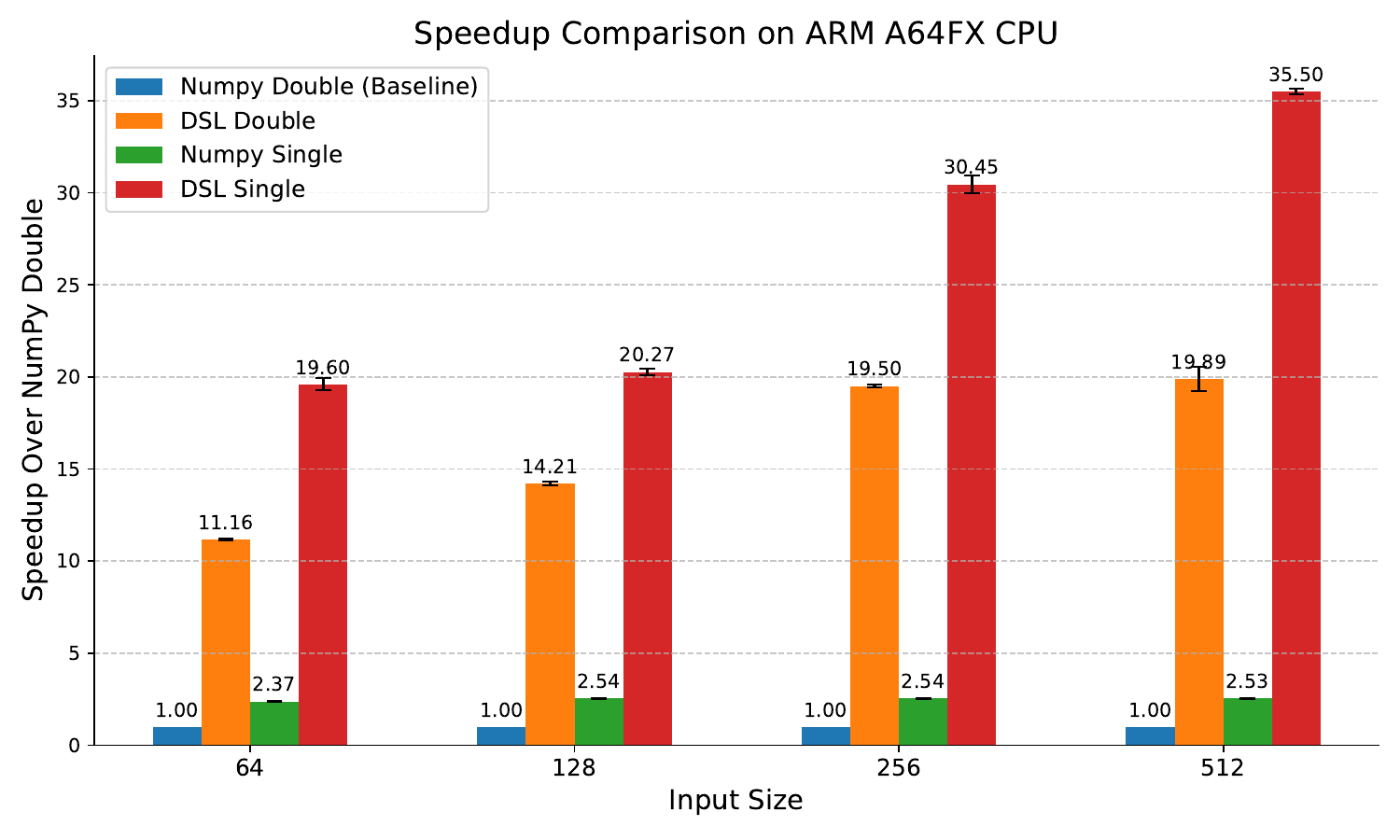}
        \caption{DSL vs. NumPy: Performance Scaling on ARM CPU.}
        \label{fig:PerfC}
    \end{subfigure}
    \hfill
    \begin{subfigure}[t]{0.48\textwidth}
        \centering
        \includegraphics[width=\textwidth]{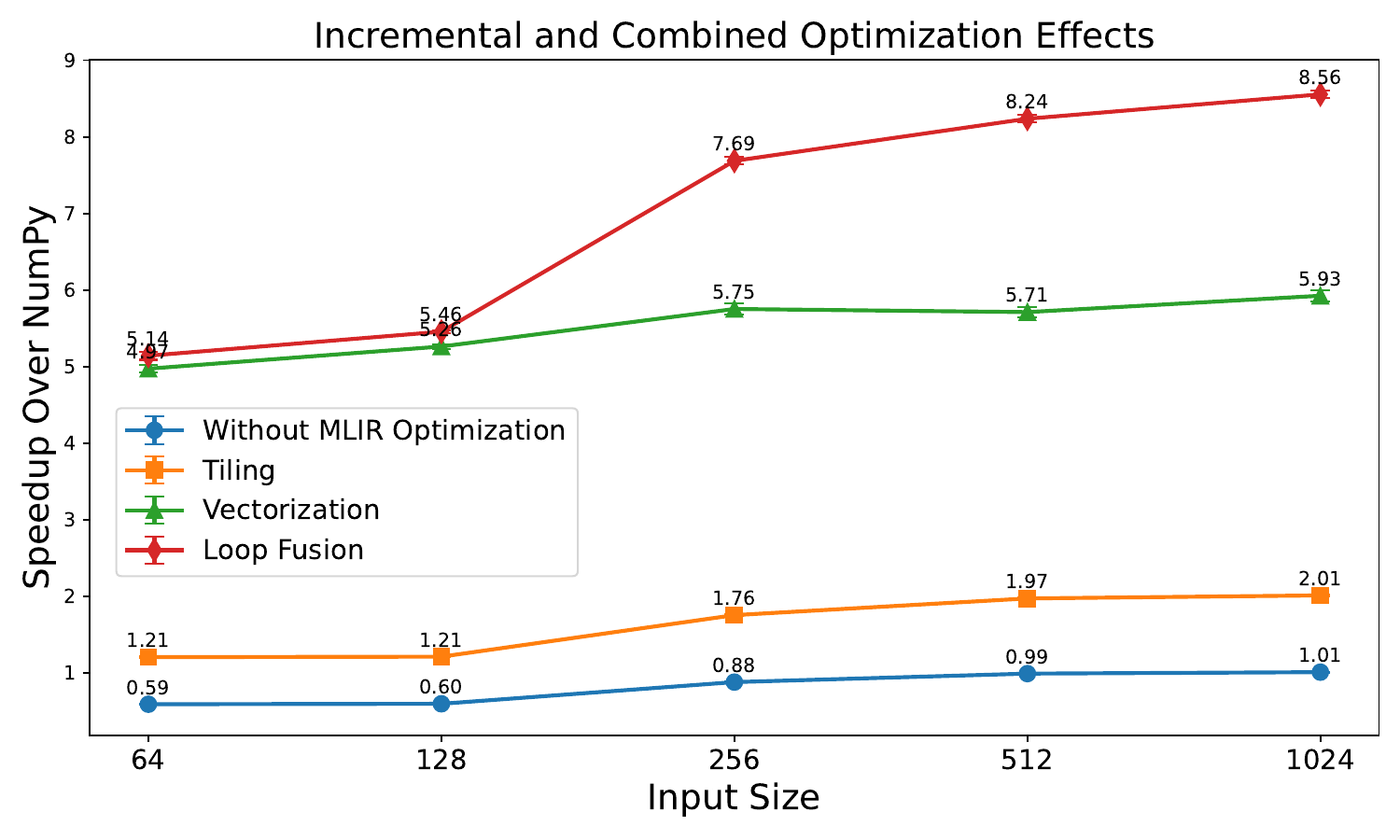}
        \caption{Performance Comparison of Different MLIR Optimization Combinations: Speedup of a Single-Precision DSL Over NumPy (Single-Precision NumPy Baseline) on Intel Architecture}
        \label{fig:PerfD}
    \end{subfigure}

    \caption{(A–C) Performance comparison of single-threaded FDTD versus NumPy (baseline: NumPy double precision) on Intel, AMD, and ARM CPUs. \\
    (D) Performance breakdown for different optimization combinations.}
    \label{fig:PerfComparison}
\vspace{-15pt}  
\end{figure*}

We first present the performance of the baseline double-precision \texttt{NumPy} implementation, comparing it against optimized MLIR implementations (both single and double precision) and a single-precision \texttt{NumPy} version for various problem sizes (from $64^3$ grids to $1,024^3$ grids), as shown in Fig.~\ref{fig:PerfComparison}. We conducted the tests on Intel, AMD, and ARM CPUs. Due to memory constraints (32 GB) on the ARM node, the NumPy implementation was unable to run the $1,024^3$ grid test; therefore, we limited the largest grid size to $512^3$ for ARM platform.

Performance is reported as speedup relative to the double-precision \texttt{NumPy} baseline for each problem size; higher values indicate better performance. 
The MLIR implementations outperform the baseline \texttt{NumPy} version across all grid sizes on the Intel platform for both single and double precision.In the single-precision case, MLIR achieves speedups ranging from $5.14\times$ for the $64^3$ grid to $8.56\times$ for the $1024^3$ grid compared to the NumPy single-precision counterpart. Overall, the speedup increases with grid size, highlighting greater optimization impact for larger problems, although we observe a plateau at very large grid sizes. Similar performance trends are also observed on AMD and ARM CPUs. An interesting observation on the ARM CPU is that, unlike other platforms, the MLIR single-precision implementation achieves less than a 2× speedup compared to double precision. Profiling data indicates this occurs due to LLVM generating a small portion of NEON instructions, despite predominantly utilizing scalable SVE instructions. This mixed vectorization approach may explain the limited performance gain. We plan to investigate and resolve this issue in future work.



As the second step of this experimental study, we assess the impact of different MLIR optimization by turning on one-by-one (Fig.~\ref{fig:PerfA}, Fig.~\ref{fig:PerfB} and Fig.~\ref{fig:PerfC} show the results for all the optimizations on). Figure~\ref{fig:PerfD} illustrates the impact of different MLIR optimizations on the Intel CPU platform. The results are broken down by different levels of optimization. The bottom line represents the basic MLIR implementation without any optimization transformations applied. Only buffer allocation is handled optimally, utilizing an efficient approach both in the OpDSL implementation and through oneshot bufferization. It can be observed that, compared to \texttt{Numpy}, this version performs slower, particularly for smaller problem sizes. 
The orange line represents the performance after enabling tiling. Here, 1D tiling was applied instead of 3D tiling, as the latter introduced significant performance overhead due to instruction cache pressure and register spilling. We observe a nearly uniform speedup across different sizes compared to the non-tiling version. The green line represents the performance with vectorization enabled, along with a series of post-vectorization optimizations applied to reduce vectorization overhead and produce a more efficient vectorized form. A significant speedup is observed across all sizes compared to the non-vectorized version. At the top, the performance of the implementation with loop fusion is shown. While a noticeable speedup is observed for larger sizes, the impact is minimal for smaller sizes such as 64 and 128.

As a final step, we profile the FDTD kernel using the \texttt{perf} tool on the Intel CPU platform, analyzing performance counters for various MLIR optimizations and the baseline \texttt{NumPy} implementation, as summarized in Table~\ref{table:Perf_Opts}. The "Vectorization Ratio \& Type" indicates the proportion of floating-point arithmetic instructions executed in vectorized versus scalar form, along with the specific SIMD instruction set utilized (e.g., AVX-512 or AVX-256). The "L1 Cache Loads" represents the number of loads from the L1 data cache, normalized to \texttt{Numpy}'s value as 1x, with other implementations compared relative to \texttt{Numpy}. Similarly, the "LLC" refers to loads from the last-level cache, also normalized to \texttt{Numpy}'s value. The "Speedup" indicates the execution time normalized to \texttt{Numpy}, where higher values represent better performance. In Table~\ref{table:Perf_Opts}, we observe that the tiled version, compared to the non-optimized version, has significantly fewer L1 data cache loads. This indicates improved data locality and data reuse, which explains the overall speedup. The MLIR vectorization achieved a high vectorization ratio by utilizing AVX-512 instructions. In contrast, the \texttt{Numpy} implementation relies on AVX-256, which offers lower parallelism. Additionally, upon analyzing the breakdown of hotspots in the assembly code, we observed that the \texttt{Numpy} implementation faces a significant bottleneck with vector insert/extract instructions, which we assume is related to handling boundary conditions. In contrast, the MLIR vectorized version shows no such bottleneck. The primary hotspot in the MLIR version is the vector division instruction, as expected. This result highlights the advantages of an efficient operator definition, high-level optimizations on tensor abstractions, and vectorization-related optimization passes. These optimizations enable consecutive vector loads instead of costly vector shuffle instructions. The profiling results with loop fusion reveal fewer L1 data cache misses and reduced LLC loads. This demonstrates improved utilization of the L1 data cache and enhanced data reuse, particularly among vector registers, resulting in better performance for larger problem sizes.
\begin{table*}[ht]
\small 
\setlength{\tabcolsep}{3pt} 

\vspace{-0pt} 

\centering
\caption{Profiling Results of Optimization Combinations for $N=256$ on Intel CPU (Single Precision).}
\resizebox{\textwidth}{!}{%
\begin{tabular}{c p{2.5cm} 
p{1.5cm} p{2.5cm} p{1.5cm} p{1.5cm} c }
\toprule
\textbf{Workloads}  
& \textbf{Vectorization Ratio \& Type}
& \textbf{L1 Cache Loads}
& \textbf{L1 Cache Load Misses}
& \textbf{LLC Loads} 
& \textbf{LLC Load Misses}
& \textbf{Speedup} 
\\
\midrule
Numpy
& 99.7\% AVX256
& 1x
& 9.04\%
& 1x
& 47.21\%
& 1x
\\
\cmidrule{1-7}
MLIR: Fusion\&Vec\&Tiling
& 98.0\% AVX512
& 0.11x
& 4.96\%
& 0.03x
& 62.94\%
& 7.69x
\\
\cmidrule{1-7}
MLIR: Vec\&Tiling
& 98.0\% AVX512
& 0.10x
& 29.10\%
& 0.21x
& 54.57\%
& 5.75x
\\
\cmidrule{1-7}
MLIR: Tiling
& 0\% Scalar
& 1.69x
& 0.13\%
& 0.02x
& 68.85\%
& 1.76x
\\
\cmidrule{1-7}
MLIR: No-Opt
& 0\% Scalar
& 4.59x
& 0.04\%
& 0.02x
& 69.12\%
& 0.88x
\\
\bottomrule
\end{tabular}%
}
\label{table:Perf_Opts}
\vspace{-0pt} 

\end{table*}

\section{Related Work}
\noindent \textbf{FDTD Method and Electromagnetic Simulations.}  
The Finite-Difference Time-Domain (FDTD) method is widely used for solving Maxwell’s equations. Notable FDTD solvers include MEEP~\cite{oskooi2010meep}, gprMax~\cite{warren2016gprmax,warren2019cuda}, and OpenEMS~\cite{liebig2013openems}, implemented in C++ or Python with MPI, OpenMP, CUDA, and SIMD support. While these solvers optimize specific aspects, their performance is constrained by platform-specific implementations and manual tuning.  

\noindent \textbf{Optimization of FDTD Solvers.}  
FDTD optimizations focus on parallelization (OpenMP, MPI), vectorization, and GPU acceleration~\cite{su2004novel,guiffaut2001parallel,livesey2012performance}. Domain decomposition enables distributed execution, while vectorization relies on intrinsics or compiler optimizations. Higher-order schemes improve computational intensity for GPUs and FPGAs~\cite{de2010introduction,mansoori2021fpga}. However, traditional compiler frameworks struggle with domain-specific optimizations, requiring manual tuning for memory access and loop optimizations.  

\noindent \textbf{Compiler-Based Optimization for Scientific Computing.}  
Compiler-driven optimizations automate transformations for stencil-based computations, including FDTD. MLIR has been applied to stencils~\cite{essadki2023code,rodriguez2023stencil}, matrix multiplication~\cite{bondhugula2020high}, FFTs~\cite{he2022fftc,he2023leveraging,10740884}, and climate modeling~\cite{gysi2021domain}. The DaCe framework~\cite{ben2023bridging} provides similar optimizations for PDE solvers and scientific simulations~\cite{andersson2023case,ziogas2019data}. In this study, we leverage MLIR for optimizing FDTD solvers, automating memory access patterns, parallelization, and loop transformations.

\section{Discussion and Conclusion}
In this work, we designed and optimized an end-to-end compiler for a high-performance 3D FDTD solver for electromagnetic cavity simulations using MLIR and its Python bindings. Utilizing tensor abstractions specifically tailored for FDTD kernels, we demonstrated efficient and straightforward implementations of loop fusion, tiling, and vectorization, significantly improving data locality, cache performance, and computational throughput. Our MLIR-based approach achieved up to a 10× speedup compared to the baseline NumPy implementation on Intel, AMD, and ARM CPUs.


Our approach demonstrates that, given only high-level tensor language input, a domain-specific compiler can automatically generate hardware-aware, high-performance kernels across multiple architectures by efficiently performing loop tiling, fusion, and vectorization. Through progressive lowering, different hardware targets share a substantial portion of high-level, hardware-agnostic abstractions and transformations, achieving excellent portability."

This work also highlights MLIR’s potential for porting Python-based applications to achieve high performance, as its Python bindings seamlessly embed and integrate with existing Python codebases.

As future work, we plan to extend our framework with GPU backends (e.g., CUDA/ROCm) and implement automated hyperparameter tuning to optimize performance across heterogeneous architectures. 


Overall, this study demonstrated the potential of MLIR and high-level tensor abstractions as an effective compiler framework for HPC scientific applications such as FDTD. Our domain-specific compiler is particularly suitable for large-scale, time-domain electromagnetic simulations, significantly enhancing application performance through automated optimizations across diverse hardware platforms.

\bibliographystyle{ieeetr}
\bibliography{ref}

\end{document}